\documentclass[preprint]{acmart}

\AtBeginDocument{%
  \providecommand\BibTeX{{%
    \normalfont B\kern-0.5em{\scshape i\kern-0.25em b}\kern-0.8em\TeX}}}

\usepackage{soul}

\newcommand{\specialcell}[2][c]{%
	\begin{tabular}[#1]{@{}c@{}}#2\end{tabular}}

\begin{document}

\title[Are We There Yet? Evaluating State-of-the-Art Neural Network based Geoparsers Using EUPEG]{Are We There Yet? Evaluating State-of-the-Art Neural Network based Geoparsers Using EUPEG as a Benchmarking Platform}

\author{Jimin Wang}
\email{jiminwan@buffalo.edu}
\affiliation{%
  \institution{GeoAI Lab, Department of Geography}
  \streetaddress{105 Wilkeson Quad}
  \city{University at Buffalo}
  \state{New York, USA}
  \postcode{14261}
}

\author{Yingjie  Hu}
\email{yhu42@buffalo.edu}
\affiliation{%
  \institution{GeoAI Lab, Department of Geography}
  \streetaddress{105 Wilkeson Quad}
  \city{University at Buffalo}
  \country{New  York, USA}}

\renewcommand{\shortauthors}{Wang and Hu}

\begin{abstract}
Geoparsing is an important task in geographic information retrieval. A geoparsing system, known as a \textit{geoparser}, takes some texts as the input and outputs the recognized place mentions and their location coordinates. In June 2019, a geoparsing competition, \textit{Toponym Resolution in Scientific Papers}, was held as one of the SemEval 2019 tasks. The winning teams developed neural network based geoparsers that achieved outstanding performances (over 90\% precision, recall, and F1 score for toponym recognition). This exciting result brings the question ``are we there yet?'', namely have we achieved high enough performances  to possibly consider the problem of geoparsing as solved? One limitation of this competition is that the developed geoparsers were tested on only one dataset which has 45 research articles collected from the particular domain of Bio-medicine. It is known that the same geoparser can have very different performances  on different  datasets. Thus, this work performs a systematic evaluation of these state-of-the-art geoparsers using our recently developed benchmarking platform EUPEG that has eight annotated datasets, nine baseline geoparsers, and eight performance metrics. The evaluation result suggests that these new geoparsers indeed improve the performances of geoparsing on multiple datasets although some challenges remain.
\end{abstract}



\maketitle

\section{Introduction}
Geoparsing is the process of recognizing and geo-locating  location mentions from   texts. It has been widely applied to various textual data, and is an important task in geographic information retrieval \cite[]{purves2018geographic}. A geoparsing system, known as a geoparser, usually functions in two steps: toponym recognition and toponym resolution. Toponym recognition detects the place mentions in texts, while toponym resolution resolves any place name ambiguity and assigns the appropriate spatial footprint (e.g., a pair of coordinates). Many geoparsers have been developed, such as CLAVIN\footnote{\url{https://clavin.bericotechnologies.com}},  the Edinburgh Geoparser \cite[]{grover2010use}, GeoTxt \cite[]{karimzadeh2019geotxt}, and TopoCluster \cite[]{delozier2015gazetteer}. 

In June 2019, an important geoparsing competition, \textit{Toponym Resolution in Scientific Papers}, was held as the SemEval 2019 Task 12, in conjunction with the Annual Conference of the North American Chapter of the Association for Computational Linguistics. This competition attracted 29 registered teams and 8 teams eventually submitted a system run \cite[]{weissenbacher2019semeval}. The winning teams all leveraged state-of-the-art neural network based models, such as BiLSTM-CRF and deep contextualized word embeddings, to design their geoparsers. Particularly, the geoparser that won the first place, DM\_NLP \cite[]{wang2019dm_nlp}, achieved over 90\% precision, recall, and F1 score for toponym recognition. This result is exciting and brings the question ``are we there yet?" A 90\% performance is not perfect but is probably sufficient for many applications. So have we already made enough progress that we can  consider the problem of geoparsing as solved?

A major limitation of the SemEval 2019 Task 12 competition is that the submitted geoparsers were tested on a single dataset which has 45 research articles from one particular domain of Bio-medicine. Existing research has shown that the same geoparser can have very different performances when  tested on different datasets  \cite[]{gritta2018s}. Accordingly,  answering the question of whether  the problem of geoparsing can be considered as solved requires a systematic evaluation of the state-of-the-art geoparsers on multiple datasets which should ideally be in different text genres (e.g., news articles, social media posts, and other types of texts).

In a recent work, we developed an online platform called EUPEG\footnote{\url{https://geoai.geog.buffalo.edu/EUPEG}} which is an Extensible and Unified Platform for Evaluating Geoparsers \cite[]{hu2018eupeg,wang2019eupeg}. EUPEG hosts a majority of the geopasing resources reported in the literature, including eight annotated datasets, nine geoparsers, and eight evaluation metrics. In addition, the eight annotated datasets are in four different text genres which are news articles, Wikipedia articles, social media posts, and texts on Web pages. The source code of EUPEG and the related geoparsing resources are shared on GitHub\footnote{\url{https://github.com/geoai-lab/EUPEG}}.

In this paper, we systematically evaluate the top geoparsers from  SemEval Task 12  using EUPEG as a benchmarking platform. We focus on the top three end-to-end geoparsers that showed the highest performances in the competition, which are DM\_NLP  \cite{wang2019dm_nlp}, UniMelb  \cite{li2019unimelb}, and UArizona \cite{yadav2019university}. We test the performances of these three geoparsers on the datasets hosted on EUPEG, and compare their performances with the other existing geoparsers. The contributions of this paper are as follows:
\begin{itemize}
	\item We conduct a systematic evaluation experiment on three state-of-the-art geoparsers, and discuss the implications and challenges based on the experiment results.
	\item We implement the three tested geoparsers based on their papers and share the source code  at \url{https://github.com/geoai-lab/GeoAI2019Geoparser} to support future research.
\end{itemize}

\section{State-of-the-Art Geoparsers}

The top three end-to-end geoparsers from SemEval Task 12 are DM\_NLP, UniMelb, and UArizona. They are all designed as pipeline systems comprising of two independent components for toponym recognition and resolution respectively. Accordingly, we describe and compare the three geoparsers based on the two components.

\subsection{Toponym Recognition} 
All three geoparsers adopt  the general Bidirectional Long Short Term Memory (BiLSTM) model for toponym recognition. However,  their models vary in regard to the selection of word embeddings, integration of character-level embeddings, concatenation with a conditional random field layer, and mechanisms of self attention. 

\textbf{DM\_NLP: } This model, ranked as the 1st place, is built upon the character and word level BiLSTM model developed by Lample et al. \cite{lample2016neural}. The authors of DM\_NLP also tested the strategies of adding four extra linguistic features into the input layer: Part-of-Speech (POS) tags, NER labels from Stanford NER, Chunking labels, and deep contextualized word representations from the ELMo word embeddings \cite{peters2018deep},  but found that only adding ELMo produces the most performance improvement. In our implementation, we  add the ELMo word embeddings as the extra linguistic feature. The  final output layer of DM\_NLP is a CRF layer. 

\textbf{UniMelb: } This model is developed by integrating  a word-level  BiLSTM \cite{hochreiter1997long} and the self-attention mechanism \cite{NIPS2017_7181}. The authors tested both the GloVe and ELMo word embeddings, and found that the model with ELMo performed better. Thus, our implementation also uses  ELMo word embeddings. 
The final layer of UniMelb is a binary softmax classifier. 

\textbf{UArizona: } This model  is  a re-implementation of a word, character, and affix level LSTM developed by Yadav et al. \cite[]{yadav2018survey}. In this model, the input of word LSTM is a concatenation of GloVe word embeddings, char embeddings represented by the output of a char  BiLSTM, and  word affix features. The word LSTM representations are given to the final CRF layer to recognize toponyms. 

We train all three toponym recognition models using a general dataset CoNLL 2003. The hyperparameters are set as the same as what reported in their papers. We use  300-dimensional pre-trained GloVe word embeddings and 1024 dimensional pre-trained EMLo embeddings from Tensorflow Hub (\url{https://tfhub.dev/google/elmo/2}). We do not update the weights of  word embeddings during the training process. 

\subsection{Toponym Resolution}  
For toponym resolution, all three geoparsers use the same general workflow of first retrieving place candidates from the GeoNames gazetteer  and then identifying the correct place instance among the candidates. However, different techniques were used by each geoparser to identify the right place instance.

\textbf{DM\_NLP: } This model constructs four groups of features, which include name string similarity, candidate attributes,  contextual features, and mention list features. These features are  then used to train a LightGBM model for toponym resolution.

\textbf{UniMelb: } This model also constructs features, including history result in the training dataset, population, GeoNames feature codes, name similarity, and ancestor names, and trains a support vector machine (SVM) for toponym resolution.

\textbf{UArizona: } This model simply uses the population heuristic for toponym resolution. Each place name is resolved to the place instance that has the highest population in GeoNames.

There is a challenge for re-implementing these toponym resolution models, that is, both DM\_NLP and UniMelb were trained  on the specific training dataset from SemEval Task 12, which consists of 105 research articles in Bio-medicine. While this is fine and even desirable for a competition (since the testing is based on 45 research articles from the same domain),  a model trained with one specific type of texts may not generalize well to other types of texts from different domains. Though we have multiple datasets available from the EUPEG platform, training the models with any of these datasets leads to the same bias issue. Ideally, the toponym resolution models of DM\_NLP and UniMelb should be trained with a large and general dataset which has labeled place instances (note that CoNLL 2003 cannot be used for training toponym resolution models) so that the general performances of these models can be measured. However, we currently do not have access to such a dataset.  Thus, we resort to a simple but general implementation, namely using the population heuristic of UArizona for all three models. Previous research, as well as the experiment result reported by the DM\_NLP team \cite[]{wang2019dm_nlp},  has shown that population heuristic is a competent baseline  and can sometimes outperform more complex models \cite[]{weissenbacher2015knowledge,delozier2015gazetteer}. 
Nevertheless, we are aware of the limitations of this simple heuristic and will discuss them with the experiment results.

\section{Experiments and Results}
\subsection{Experiments on EUPEG}
The three neural network based geoparsers are tested on  EUPEG. As a benchmarking platform, EUPEG provides eight annotated corpora, nine geoparsers, and eight performance metrics. Table \ref{EUPEG_info}  summarizes these resources. More detailed descriptions on each of the resources can be found in our full paper about EUPEG \cite{wang2019eupeg}. We provide brief descriptions below to make this current paper self-contained.

\begin{table}[ht]
	\caption{Datasets, geoparsers, and metrics on EUPEG}
	\label{EUPEG_info}
	\vspace*{-0.2cm}
		\begin{tabular}{cl}
			\toprule
			\textbf{Category} &\textbf{Resources}\\
			\midrule
			Datasets & LGL, GeoVirus, TR-News, GeoWebNews, WikToR, \\
			&GeoCorpora, Hu2014, Ju2016\\
			Geoparsers&GeoTxt, The Edinburgh Geoparser, CLAVIN,\\
			&Yahoo! PlaceSpotter, CamCoder, TopoCluster, \\&StanfordNER+Population, SpaCyNER+Population,\\
			&
			DBpedia Spotlight\\
			Metrics  & Precision, Recall, F1 score, Accuracy, Mean,\\
			& Median, AUC, Accuracy@161\\
			\bottomrule
	\end{tabular} 
\end{table}

The eight datasets are in four different text genres: news articles, Wikipedia articles,  social  media  posts, and  Web  pages. Particularly, \textit{LGL}, \textit{GeoVirus}, \textit{TR-News}, and \textit{GeoWebNews} contain annotated news articles; \textit{WikToR} is a Wikipedia dataset; \textit{GeoCorpora} is a social media dataset that contains annotated tweets; and \textit{hu2014} and \textit{Ju2016} are two corpora that contain texts retrieved from Web pages. These diverse datasets enable a more comprehensive evaluation on the performance of a geoparser. It is worth noting that these datasets were annotated  by researchers from different domains (e.g., geography, linguistics, and computer science). As a result, there exist  differences in the words and phrases that are considered  as toponyms. All datasets annotate administrative units, such as cities, towns, and countries. However, some datasets, such as \textit{LGL} and \textit{GeoWebNews}, also consider demonyms (e.g., Canadian) as toponyms. The toponyms in the dataset \textit{GeoCorpora}, in addition to administrative units, also include natural features (e.g., lakes and mountains) and facilities (e.g., streets and buildings) which are not included in some other datasets (e.g., \textit{GeoVirus}). This definition difference of toponyms directly affects the performances  of the same geoparser on different datasets.

The nine geoparsers hosted on EUPEG use a variety of heuristics and machine learning based methods. Particularly, \textit{GeoTxt}, \textit{The Edinburgh Geoparser}, and \textit{CLAVIN} use a named entity recognition tool for toponym recognition and a number of heuristics (e.g., the level of an administrative unit and population) for toponym resolution. \textit{TopoCluster} uses Stanford NER for toponym recognition and generates geographic profiles of words for toponym resolution. \textit{CamCoder} is a deep learning based geoparser that leverages a Convolutional Neural Network (CNNs) model. \textit{Yahoo! PlaceSpotter} is an industrial geoparser which provides an
online REST API (at the time of writing this paper, the online service of \textit{Yahoo! PlaceSpotter} has become unavailable). In addition to the six geoparsers, EUPEG also includes two baseline geoparsers that are developed using Stanford NER and SpaCy NER with a population heuristic, as well as \textit{DBpedia Spotlight}, a general named entity recognition and linking (NERL) tool that can be used as a geoparser.

The eight performance metrics provided on EUPEG include standard metrics from information retrieval as well as geographic distance based metrics designed for measuring the quality of the resolved geographic locations. The metrics of \textit{precision}, \textit{recall}, \textit{F1 score} and \textit{accuracy} evaluate the ability of a geoparser in correctly recognizing toponyms from texts. Particularly, the metric of \textit{accuracy} is used in situations when only some of the mentioned toponyms are annotated. The metrics of \textit{mean} and \textit{median} measures how far  the resolved location is away from the ground-truth location (in kilometers). The metric of \textit{accuracy@161} measures the percentage of the resolved locations that are within 161 kilometers (100 miles) of the ground truth. The metric of \textit{AUC} (Area Under the Curve) measures a normalized distance error by calculating the area under a distance error curve.

The three neural network based geoparsers from SemEval Task 12 are tested using the datasets  from EUPEG. We quantify their performances using the discussed metrics, and compare their performances with those of the other  geoparsers hosted on EUPEG.


\subsection{Results}
The experiment results contain the performances of the three state-of-the-art  geoparsers on the eight datasets in comparison with the other existing geoparsers. In the following, we present and discuss the experiment results on three datasets, namely \textit{GeoVirus}, \textit{GeoCorpora}, and \textit{Ju2016}. We provide the results on the other five datasets in Appendix A. 


\subsubsection{Results on GeoVirus.} GeoVirus is a corpus that contains 229 news articles. This dataset was originally developed by Gritta et al. \cite{gritta2018melbourne}, and the news articles were collected during 08/2017 - 09/2017, covering the topics about global disease outbreaks and epidemics. GeoVirus is a relatively easy dataset since most location mentions refer to  prominent place instances (e.g., major cities or countries) and the texts  from news articles  are well formatted. The evaluation results on GeoVirus are summarized in Table \ref{GeoVirusTable}.  Since the online service of Yahoo! PlaceSpotter has become unavailable, its performance is not included in the experiment results. 

\begin{table}[H]
	\caption{Evaluation results on GeoVirus}
	\label{GeoVirusTable}
	\vspace*{-0.2cm}
		\begin{tabular}{ccccccccc}
			\toprule
			Geoparser&precision&recall&f\_score&mean &median &acc@161&AUC\\
			\midrule
			\specialcell{DM\_NLP+Pop}&0.917&0.916&\textbf{0.917}&770.337&48.676&0.655&0.378\\
			StanfordNER&0.927&0.903&0.915&791.296&48.676&0.655&0.378\\
			\specialcell{UniMelb+Pop}&0.882&\textbf{0.936}&0.908&777.234&48.466&0.657&0.379\\
			\specialcell{UArizona }&0.887&0.859&0.873&769.810&55.635&0.640&0.386\\			
			CamCoder&\textbf{0.940}&0.802&0.866&619.397&33.945&0.770&0.336\\
			TopoCluster&0.877&0.813&0.844&599.632&63.858&0.673&0.407\\
			GeoTxt&0.857&0.726&0.786&487.874&36.255&0.787&0.338\\
			CLAVIN&0.913&0.637&0.750&522.176&35.503&0.786&0.320\\
			DBpedia &0.792&0.616&0.693&1272.937&122.314&0.533&0.406\\
			Edinburgh &0.860&0.559&0.678&\textbf{435.799}&\textbf{33.187}&\textbf{0.807}&\textbf{0.319}\\
			SpaCyNER&0.721&0.382&0.499&788.231&40.653&0.698&0.367\\
			\bottomrule
	\end{tabular} 
\end{table}

The geoparsers in the table above are ordered by their F1 scores. The metrics of \textit{precision}, \textit{recall}, and \textit{f\_score} evaluate the performances of a geoparser for  toponym recognition. The other four metrics evaluate the performance of a geoparser in resolving a toponym to its correct geographic location. 
It can be seen that the three top geoparsers from SemEval Task 12 indeed rank very high based on their F1 scores for the task of toponym recognition. However, the off-the-shelf StanfordNER also shows very competitive performance on this simple dataset. In terms of toponym resolution, \textit{The Edinburgh Geoparser} performs the best, although the median error distance for most geoparsers are within 100 km. Since most place mentions refer to their prominent instances, the population heuristic works well. It is worth noting that toponym resolution is performed based on only the  toponyms recognized in the previous step. Thus, the metrics of \textit{mean}, \textit{median}, \textit{acc@161}, and \textit{AUC} are measured based on different numbers of toponyms that need to be resolved.

\subsubsection{Results on GeoCorpora} GeoCorpora is a social media  corpus that contains annotated tweets. GeoCorpora was developed by Wallgr{\"u}n et al. \cite{wallgrun2018geocorpora}, and their  original  paper reported 2,287 annotated tweets. Due to deletions, only 1,639 tweets are recovered on EUPEG. Compared to GeoVirus, GeoCorpora has two unique characteristics. First, the texts in GeoCorpora are short sentences (tweets within 140 characters) which provide only limited contextual information. Second, the content of tweets does not strictly follow grammatical rules and often contains abbreviations. Accordingly, GeoCorpora presents a more difficult dataset than GeoVirus. The evaluation results are summarized in Table \ref{GeoCorporaTable}.

\begin{table}[H]
	\caption{Evaluation results on GeoCorpora}
	\label{GeoCorporaTable}
	\vspace*{-0.2cm}
		\begin{tabular}{cccccccc}
			\toprule
			Geoparser&precision&recall&f\_score&mean&median&acc@161&AUC\\
			\midrule
			\specialcell{DM\_NLP+Pop}&0.888&\textbf{0.669}&\textbf{0.763}&1249.865&0.000&0.661&0.288\\
			\specialcell{UniMelb+Pop}&0.852&0.661&0.745&1245.992&0.000&0.659&0.289\\
			\specialcell{UArizona}&0.892&0.598&0.716&1079.012&0.000&0.668&0.278\\		
			GeoTxt&\textbf{0.926}&0.521&0.667&714.94&0.000&0.876&0.116\\
			StanfordNER&0.899&0.526&0.664&1063.473&0.000&0.676&0.270\\
			CamCoder&0.904&0.503&0.647&1024.723&0.000&0.820&0.163\\		
			TopoCluster&0.882&0.506&0.643&575.225&32.948&0.698&0.361\\
			DBpedia &0.865&0.500&0.633&669.105&33.816&0.654&0.352\\
			Edinburgh &0.832&0.505&0.628&958.401&0.000&0.848&0.139\\		
			SpacyNER&0.705&0.467&0.562&982.137&0.000&0.752&0.224\\
			CLAVIN&0.907&0.341&0.496&\textbf{373.563}&0.000&\textbf{0.913}&\textbf{0.084}\\
			\bottomrule
	\end{tabular}
\end{table}

As can be seen, the F1 scores of all three geoparsers drop considerably on this more difficult dataset. However, it is worth emphasizing that  DM\_NLP increases the best possible F1 score  from about 0.66 (by GeoTxt) to about 0.76, which is a large improvement. For toponym resolution, population heuristic is still a relatively effective approach on this dataset based on the zero median error distances achieved by the three new geoparsers. However, population heuristic is not as effective as other models, such as CLAVIN, GeoTxt, Edinburgh, and CamCoder (based on their higher acc@161 and lower AUC). Again, when we interpret the values of \textit{mean}, \textit{median}, \textit{acc@161} and \textit{AUC}, it is necessary to take into account the factor that toponym resolution is evaluated based on  the different numbers of recognized toponyms from the previous step.

\subsubsection{Results on Ju2016} Ju2016 is a corpus containing short sentences retrieved from various Web pages. This dataset was created by Ju et al. \cite[]{ju2016things} who developed a script using Microsoft Bing Search API to automatically retrieve sentences containing highly ambiguous US place names (e.g., ``Washington"). This corpus contains 5,441 entries in total and the average length of each entry is 21 words. This is a very difficult dataset, because the sentences are short (limited contextual information), place names are ambiguous, and upper and lower cases are not differentiated (all words are converted to lower case). Since this is an automatically created dataset, not all place mentions are annotated and as a result, precision, recall, and F1 score cannot be used as performance metrics. Following previous research \cite[]{gritta2018s},  we use  \textit{accuracy} which measures the percentage of place names that are correctly recognized among all annotated place names. The  results on Ju2016 are provided in Table \ref{Ju2016Table}.

\begin{table}[ht]
	\caption{Evaluation results on Ju2016}
	\label{Ju2016Table}
	\vspace*{-0.2cm}
		\begin{tabular}{cccccc}
			\toprule
			Geoparser&accuracy&mean&median&acc@161&AUC\\
			\midrule
			GeoTxt&\textbf{0.463}&2609.734&1616.741&0.032&0.731\\
			DBpedia &0.447&3101.087&1417.795&\textbf{0.111}&\textbf{0.698}\\
				\specialcell{UniMelb+Pop}&0.379&3301.993&2081.599&0.020&0.758\\
			TopoCluster&0.158&4026.270&1547.266&0.036&0.752\\
			\specialcell{DM\_NLP+Pop}&0.097&3357.802&2266.718&0.020&0.760\\
			\specialcell{UArizona}&0.036&2433.890&1966.937&0.029&0.739\\			
			StanfordNER&0.01&2027.016&2459.841&0&0.745\\
			CamCoder&0.004&\textbf{1559.437}&\textbf{1389.716}&0.042&0.709\\
			SpacyNER&0.004&3330.696&3478.187&0&0.818\\
			CLAVIN&0&0&0&0&0\\
			Edinburgh &0&0&0&0&0\\
			\bottomrule
	\end{tabular}
\end{table}

As can be seen, many geoparsers show dramatically decreasing performances on this very difficult dataset. Two geoparsers,  CLAVIN and Edinburgh, completely fail on this dataset  which does not have word capitalization. Many other geoparsers,  including DM\_NLP and UArizona, also largely fail on this dataset due to their use of case-sensitive features, such as separate character-level embeddings for upper and lower case characters.
UniMelb is an exception among the three  geoparsers that performs still relatively well. Its performance can be attributed to its model design that does not include case sensitive character-level embeddings as DM\_NLP and UArizona do.
The highest accuracy is achieved by   GeoTxt and DBpedia Spotlight, but all geoparsers show very low performances for toponym resolution based on the low acc@161 and high AUC scores.   \textit{Ju2016} is an artificially created dataset whose difficulty was deliberately increased for the purpose of testing geoparsers. It is less likely for a real world corpus to contain so many different  place instances all sharing the same name (e.g., the many ``Washington"s in this dataset). However, many real world corpora are likely to have irregular case alternations, and a robust geoparser should be able to accommodate such variations.


\subsection{Discussion}
So are we there yet? Have we achieved sufficient progress on geoparsing to possibly consider the problem as solved? In our view, the answer is ``it depends". It depends on the characteristics of the textual corpus on which geoparsing is performed. If the dataset contains well-formatted articles and is mostly about prominent places throughout the world (e.g., international news articles), then the answer is probably ``yes" since the state-of-the-art geoparser, DM\_NLP can achieve over 0.91 in precision, recall, and F1 score, and a relatively low toponym resolution error using a simple population heuristic. In fact, for such a dataset, one can even use the off-the-shelf StanfordNER combined with a population heuristic, saving the time  for training a complex deep neural network model. On the other hand, if the dataset contains mostly short and informally-written sentences with ambiguous place names, then the answer is ``no" since many of our current geoparsers will largely fail on such a dataset. In addition to handling toponym ambiguity, typos, name variations, case alterations, and limited contexts in short texts, future geoparsing research could also explore a number of directions, which are discussed as follows.  

\textit{Geoparsing without population information.} As shown in our experiment results, an off-the-shelf NER tool combined with a simple population heuristic can already provide competent performance for geoparsing. However, there are situations in which population information is not available in the gazetteer, or the toponyms to be parsed do not have population (e.g., toponyms about streets or mountains). Methods that do not rely on population information need to be employed in these situations. For example, Moncla et al. \cite{moncla2014geocoding} leveraged clustering techniques to disambiguate toponyms contained in a hiking description corpus.

\textit{Geoparsing fine-grained locations.} A majority of geoparsing research so far has focused on recognizing and resolving toponyms at a geographic level higher than cities, towns, and villages. Sometimes, we may want to geoparse fine-grained locations within a city, such as street names, or the names of parks and monuments. A geoparser based on a large and general gazetteer will not be able to geo-locate such fine-grained locations. In a recent work,  Alex et al.   adapted the Edinburgh Geoparser to process literary text containing fine-grained place names located in and around  the City of Edinburgh, and also released a non-copyrighted gold standard datasets to support research in this direction \cite{alexgeoparsing}.

\textit{Geoparsing with gazetteers beyond GeoNames.} Gazetteer plays a critical role in linking recognized toponyms and their geographic locations. However, most existing geoparsers only use GeoNames as their gazetteer. This, to some extent, can be attributed to the fact that many corpora are annotated based on GeoNames, and as a result, geoparsers are also developed based on GeoNames for evaluation convenience. As discussed in the previous point, a geoparser based on GeoNames will not be able to parse fine-grained place names. Besides, such a geoparser cannot process the historical texts in the context of digital humanities applications. An ideal geoparser, therefore, should allow users to switch the underlying gazetteer to one beyond GeoNames.

\section{Conclusion and Future Work}

 
Geoparsing is an important research problem. This  paper presents our work on evaluating the three state-of-the-art geoparsers coming out from the SemEval-2019 Task 12 competition in June 2019. This work is motivated by the outstanding performances of these geoparsers in the competition. As a result, we set out to examine whether we have made enough progress to possibly consider the problem of geoparsing as solved. We systematically tested the top three geoparsers on our benchmarking platform EUPEG. The results suggest that these new geoparsers indeed improve the highest possible scores on multiple datasets, and the problem of geoparsing well-formatted texts referring to prominent place instances could be considered as solved. Meanwhile, some challenges remain, such as  geoparsing toponyms from informally-written texts with ambiguous place names.

This work can be extended in several directions. As discussed previously, we used a simple population heuristic for the toponym resolution component of the three geoparsers. Therefore, a next step is to develop a general toponym resolution dataset and use it to train the machine learning models described in the papers of DM\_NLP and UniMelb. Second, EUPEG currently does not contain historical corpora. As a result, it cannot be used for testing the performances of geoparsers on historical texts for humanities applications. An extension of EUPEG with historical corpora (e.g., 19th century newspapers and fictional works) can make this platform even more useful for researchers in digital humanities. A similar idea can be applied to extending EUPEG with non-English corpora.  Third, EUPEG currently evaluates only end-to-end geoparsers, and it could be useful to extend EUPEG with the capability of evaluating software tools designed for toponym recognition or resolution only. We have shared the source code of EUPEG, along with the datasets under open licenses, on GitHub at: \url{https://github.com/geoai-lab/EUPEG}. The source code of the three implemented neural network geoparsers tested in this work is also shared on GitHub at: \url{https://github.com/geoai-lab/GeoAI2019Geoparser}.  We hope that these resources can help support the future work of the community to further advance geoparsing.





\section*{Acknowledgments}
The authors would like to thank the four anonymous reviewers for their constructive comments and suggestions.

\bibliographystyle{ACM-Reference-Format}
\bibliography{sample-base}

\section*{Appendix A: Experiment Results on the Other Five Datasets}

\begin{table}[H]
	\caption{Evaluation results on LGL}
	\label{LGLTable}
		\begin{tabular}{clllllll}
			\toprule
			Geoparser&precision&recall&f\_score&mean (km)&median (km)&acc@161&AUC\\
			\midrule
			DBpedia Spotlight&\textbf{0.813}&\textbf{0.635}&\textbf{0.713}&1465.669&7.953&0.643&0.361\\
			\specialcell{DM\_NLP+Pop}&0.730&0.630&0.677&1517.406&12.852&0.582&0.373\\
			StanfordNER&0.744&0.622&0.677&1485.954&11.56&0.59&0.367\\
			\specialcell{UniMelb+Pop}&0.694&0.653&0.673&1527.597&13.340&0.581&0.375\\
			TopoCluster&0.763&0.577&0.657&1209.39&18.959&0.625&0.379\\
			CamCoder&0.811&0.548&0.654&837.126&0.02&0.717&0.248\\
			\specialcell{UArizona}&0.717&0.533&0.611&1570.982&13.477&0.582&0.372\\			
			GeoTxt&0.747&0.503&0.601&1544.283&0.044&0.633&0.312\\
			CLAVIN&0.808&0.444&0.573&1261.408&0.012&0.701&0.262\\
			Edinburgh Geoparser&0.723&0.383&0.501&\textbf{611.06}&\textbf{0.005}&\textbf{0.819}&\textbf{0.172}\\
			SpacyNER&0.493&0.371&0.423&1702.214&7.685&0.561&0.381\\
			\bottomrule
	\end{tabular}
\end{table}

\begin{table}[H]
	\caption{Evaluation results on TR-News}
	\label{TR-NewsTable}
		\begin{tabular}{clllllll}
			\toprule
			Geoparser&precision&recall&f\_score&mean (km)&median (km)&acc@161&AUC\\
			\midrule
			\textbf{TopoCluster}& 0.883&\textbf{0.714}&\textbf{0.79}&1140.551&29.336&0.623&0.387\\
			CamCoder&0.897&0.638&0.746&863.856&0&0.824&0.161\\
			StanfordNER&0.89&0.731&0.803&1170.53&0&0.711&0.261\\
			DBpedia Spotlight&0.861&0.631&0.728&1702.697&115.055&0.53&0.434\\
			\specialcell{UniMelb+Pop}&0.842&0.621&0.715&1151.496&0.000&0.729&0.244\\
			\specialcell{UArizona}&0.871&0.580&0.696&1217.766&0.000&0.697&0.265\\			
			GeoTxt&0.824&0.596&0.692&1017.3&0&0.801&0.168\\
			\specialcell{DM\_NLP+Pop}&0.749&0.618&0.677&1313.811&0.000&0.688&0.280\\
			CLAVIN&\textbf{0.908}&0.505&0.649&955.49&0&0.829&0.149\\
			Edinburgh Geoparser&0.709&0.538&0.612&\textbf{770.227}&0&\textbf{0.85}&\textbf{0.127}\\
			SpacyNER&0.659&0.402&0.5&1249.594&0&0.739&0.239\\
			\bottomrule
	\end{tabular}
\end{table}

\begin{table}[H]
	\caption{Evaluation results on GeoWebNews}
	\label{GeoWebNewsTable}
		\begin{tabular}{clllllll}
			\toprule
			Geoparser&precision&recall&f\_score&mean (km)&median (km)&acc@161&AUC\\
			\midrule
			StanfordNER&0.885&\textbf{0.635}&\textbf{0.739}&818.282&0&0.698&0.257\\
			\specialcell{UniMelb+Pop}&0.851&0.628&0.722&877.210&0.003&0.691&0.268\\
			\specialcell{DM\_NLP+Pop}&0.865&0.612&0.717&866.754&0.003&0.697&0.265\\
			CamCoder&0.895&0.562&0.691&723.122&0&0.839&0.15\\
			TopoCluster&0.838&0.559&0.67&597.082&42.46&0.68&0.357\\
			Edinburgh Geoparser&0.819&0.538&0.65&346.873&0&0.921&0.071\\
			DBpedia Spotlight&0.847&0.51&0.637&736.677&94.298&0.564&0.396\\
			GeoTxt&0.771&0.479&0.591&421.073&0&0.903&0.086\\
			CLAVIN&\textbf{0.909}&0.394&0.549&\textbf{210.905}&0&\textbf{0.937}&\textbf{0.06}\\
			SpacyNER&0.784&0.415&0.543&1053.063&55.555&0.661&0.396\\
			\specialcell{UArizona}&0.860&0.357&0.504&928.186&1.046&0.648&0.290\\
			\bottomrule
	\end{tabular}
\end{table}

\begin{table}[H]
	\caption{Evaluation results on Hu2014}
	\label{Hu2014Table}
		\begin{tabular}{clllllll}
			\toprule
			Geoparser&accuracy&mean (km)&median (km)&acc@161&AUC\\
			\midrule
			GeoTxt&\textbf{0.85}&928.839&1074.851&0.044&0.653\\
			\specialcell{UArizona}&0.813&2353.558&2575.671&0.000&0.763\\	
			TopoCluster&0.794&926.444&1074.851&0.008&0.674\\		
			StanfordNER&0.787&2277.44&2575.671&0&0.759\\
			\specialcell{DM\_NLP+Pop}&0.700&2285.899&2575.671&0.000&0.759\\
			DBpedia Spotlight&0.688&8846.334&10154.526&0.018&0.883\\
			\specialcell{UniMelb+Pop}&0.681&11007.875&11040.163&0.000&0.939\\
			SpacyNER&0.681&2322.062&2575.671&0&0.762\\
			Edinburgh Geoparser&0.656&\textbf{854.222}&1030.441&0.114&0.607\\
			CLAVIN&0.65&951.498&1074.851&0.048&0.653\\
			CamCoder&0.637&1250.964&\textbf{655.799}&\textbf{0.294}&\textbf{0.536}\\
			\bottomrule
	\end{tabular}
\end{table}

\begin{table}[H]
	\caption{Evaluation results on WikToR}
	\label{WikToRTable}
		\begin{tabular}{clllllll}
			\toprule
			Geoparser&accuracy&mean (km)&median (km)&acc@161&AUC\\
			\midrule
			\specialcell{UniMelb+Pop}&\textbf{0.681}&4775.239&2804.149&0.171&0.712\\
			\specialcell{DM\_NLP+Pop}&0.673&4842.807&2882.810&0.167&0.715\\
			DBpedia Spotlight&0.604&2272.737&5.226&0.545&0.391\\
			StanfordNER&0.54&4602.864&2513.181&0.184&0.702\\
			SpacyNER&0.518&4785.691&2917.174&0.157&0.72\\
			GeoTxt&0.506&4706.664&2644.041&0.179&0.701\\
			TopoCluster&0.47&3800.378&1531.454&0.26&0.628\\
			CamCoder&0.424&\textbf{1150.051}&33.967&0.588&\textbf{0.37}\\
			\specialcell{UArizona}&0.383&4940.599&3172.620&0.148&0.730\\			
			Edinburgh Geoparser&0.298&2165.389&\textbf{3.49}&\textbf{0.591}&0.378\\
			CLAVIN&0.215&4220.027&2331.119&0.154&0.702\\
			\bottomrule
	\end{tabular}
\end{table}

\end{document}